\documentclass[conference,10pt]{IEEEtran}
\IEEEoverridecommandlockouts
\usepackage{tabularx}
\usepackage{lscape}
\usepackage{mathtools}
\usepackage{booktabs}

\DeclareMathOperator*{\argmin}{arg\,min}

\usepackage{cite}
\usepackage{amsmath,amssymb,amsfonts}
\usepackage{algorithmic}
\usepackage{graphicx}
\usepackage{textcomp}
\usepackage{xcolor}
\usepackage{enumitem}
\usepackage[square,numbers]{natbib}

\usepackage[ruled,vlined, linesnumbered]{algorithm2e}

\usepackage[acronym]{glossaries} % Acronyms
\newacronym{dl}{DL}{Deep Learning}
\newacronym{snns}{SNNs}{Spiking Neural Networks}
\newacronym{dnns}{DNNs}{Deep Neural Networks}
\newacronym{cnn}{CNN}{Convolutional Neural Network}
\usepackage{hyperref}
\usepackage{tikz}
\newcommand*\circled[1]{\tikz[baseline=(char.base)]{
            \node[shape=circle,draw,inner sep=0.8pt, minimum size=2pt] (char) {#1};}}
            
\newcommand{\rpoint}[1]{\circled{{\fontfamily{pcr}\selectfont\footnotesize{#1}}}}

\usepackage[a4paper, total={184mm,239mm}]{geometry}
\def\BibTeX{{\rm B\kern-.05em{\sc i\kern-.025em b}\kern-.08em
    T\kern-.1667em\lower.7ex\hbox{E}\kern-.125emX}}

\usepackage{fancyhdr,lipsum}
\fancypagestyle{firstpage}{% Page style for first page
  \fancyhf{}% Clear header/footer
  \fancyhead[C]{To appear at the 24th Design, Automation and Test in Europe (DATE'21), February, 2021.}% Header
  \fancyfoot[C]{\thepage}% Footer
}

\begin{document}

\title{Securing Deep Spiking Neural Networks against Adversarial Attacks through Inherent Structural Parameters}

\author{\IEEEauthorblockN{Rida El-Allami$^{1,*}$\thanks{*These authors contributed equally to this work.}, Alberto Marchisio$^{2,*}$, Muhammad Shafique$^3$, Ihsen Alouani$^1$}
%\IEEEauthorblockA{\textit{$^1$Politecnico di Torino, Turin, Italy}} 
%\IEEEauthorblockA{\textit{$^2$Technische Universität Wien, Vienna, Austria}}\\
\IEEEauthorblockA{
\textit{$^1$ IEMN CNRS-UMR8520, Université Polytechnique Hauts-De-France, Valenciennes, France}\\
\textit{$^2$Institute of Computer Engineering, Technische Universität Wien, Vienna, Austria}\\
\textit{$^3$Division of Engineering, New York University Abu Dhabi, UAE}\\
\textit{Email: rida.elallami@etu.uphf.fr, alberto.marchisio@tuwien.ac.at, muhammad.shafique@nyu.edu, ihsen.alouani@uphf.fr
}}\vspace*{-20pt}}

%\title{A Thorough Study on Deep Spiking Neural Networks Inherent Security}

%\title{DNN vs. SNN: A Methodology to Perform a Comparative Study from a Security Perspective}

% \and
% \and
% \and
% \and
% \and

\maketitle
\thispagestyle{firstpage}

%\color{red}Authors:\\Rida El Allami, Alberto Marchisio, Ihsen Alouani, Muhammad Shafique

% Tracks:
% \begin{itemize}
%     \item DT6 Design and Test of Secure Systems
% \end{itemize}
% \normalcolor

\begin{abstract}

\gls{dl} algorithms have gained popularity owing to their practical problem-solving capacity. However, they suffer from a serious integrity threat, i.e., their vulnerability to adversarial attacks. In the quest for \gls{dl} trustworthiness, recent works claimed the inherent robustness of \gls{snns} to these attacks, without considering the variability in their structural spiking parameters. 
This paper explores the security enhancement of \gls{snns} through internal structural parameters. Specifically, we investigate the \gls{snns} robustness to adversarial attacks with different values of the neuron's firing voltage thresholds and time window boundaries. We thoroughly study \gls{snns} security under different adversarial attacks in the strong white-box setting, with different noise budgets and under variable spiking parameters. Our results show a significant impact of the structural parameters on the \gls{snns}' security, and promising sweet spots can be reached to design trustworthy \gls{snns} with $85\%$ higher robustness than a traditional non-spiking \gls{dl} system. 
To the best of our knowledge, this is the first work that investigates the impact of structural parameters on \gls{snns} robustness to adversarial attacks. The proposed contributions and the experimental framework is available \href{https://github.com/rda-ela/SNN-Adversarial-Attacks}{online}\footnote{\url{https://github.com/rda-ela/SNN-Adversarial-Attacks}} to the community for reproducible research. 
%Recently, Spiking Neural Networks (SNNs) emerged as a promising candidate to solving the DNNs high power consumption challenge, especially in the context of embedded systems and the Edge. 
%: (1) Robustness, especially under adversarial attack scenarios, and (2) the high power and resource consumption. Due to their high power efficiency, Spiking Neural Networks (SNNs) are promising candidates to solving the second challenge. However, adversarial attacks remain a serious thereat to neural networks integrity. 

%, they suffer from two main problems: (i) adversarial attacks, i.e., carefully crafted noise added to the input that forces the DNN's output to wrong labels; and (ii) 

%Spiking Neural Network (SNN) as a potential candidate for inherent robustness against adversarial attacks.
%\begin{itemize}
    % \item Importance of ML
    % \item Why SNNs are so promising? 
    % \item Threat to ML security; Adversarial attacks
    % \item Intuition on SNNs flexibility, allowing higher exploring dimensions. Not yet fully explored in literature. 
 %   \item We extensively explored inherent robustness to adversarial attacks. We found that compared to CNNs, SNNs are more secure. Moreover, we are first to explore the impact of structural parameters, i.e., firing threshold voltage and neurons' time window on SNNs robustness. Sweet spots have been found. We will open-source the complete code of our experiments and the robust SNN models at https://BlindedLink.
%\end{itemize}
\end{abstract}

\begin{IEEEkeywords}
SNN, Spiking Neural Networks, Security, Machine Learning, Deep Learning, Neuromorphic, Adversarial Attacks, Robustness, Parameters, Optimization, Analysis.
\end{IEEEkeywords}

\section{Introduction} \label{sec:intro}
%\ihsen{@Alberto: Could you please take care of the introduction?}
The recent advances in \gls{dnns} have made them the de-facto standard algorithm for several Machine Learning (ML) applications in sectors such as finance, robotics, healthcare and computer vision~\cite{Capra2020Updated}. 
However, the trustworthiness of \gls{dnns} is threathened by adversarial attacks~\cite{vulnerable, RobustML_shafique, RobustML2_shafique}. A malicious actor is able to jeopardize DNNs integrity with a small and imperceptible input perturbation. In safety-critical applications (e.g., automotive~\cite{Luckow2016DLAutomotive}, medicine~\cite{Jang2019DLmedical}, banking~\cite{Ozbayoglu2020DLFinance}), these attacks can cause catastrophic consequences and considerable losses. For example, misclassifying a \emph{stop} traffic sign as a \emph{speed limit} sign could lead to material and human damages. Another scenario is in financial transactions and automatic bank check processing: automatic handwritten character reading of bank checks~\cite{cheque}. An attacker could easily fool the model to predict wrong bank account numbers or wrong amount of money.% A dangerous situation, especially with such large sums of money at stake.

On the other hand, Spiking Neural Networks (\gls{snns}) recently emerged as an attractive alternative, due to their biological plausibility and similarity to the human brain's functionality~\cite{Kasinski2011IntroSNN}. Moreover, neuromorphic hardware can exploit the asynchronous communication between neurons and the event-based propagation of the information through layers to achieve high energy-efficiency~\cite{Capra2020SurveyDNN, Putra2020FSpiNN}. These characteristics led to an increasing interest in developing neuromorphic architectures such as IBM TrueNorth~\cite{Merolla2014Truenorth} and Intel Loihi~\cite{Davies2018Loihi}. More interestingly from a security perspective, recent studies claimed inherent robustness of SNNs compared to traditional DNNs~\cite{Marchisio2019SNNAttack, Sharmin2019ACA}. 

\subsection{Target Research Problem and Research Challenges}

While several systematic analyses of the reliability and security of traditional (i.e., non-spiking) \gls{dnns} against adversarial attacks have been thoroughly conducted~\cite{RobustML_shafique, RobustML2_shafique, D&T, Hanif2019SalvageDNN, Hoang2020FT-ClipAct, Khalid2019FAdeML, Khalid2019QuSecNets}, the corresponding analysis of SNNs robustness is still under-explored. Due to their bio-inspired aspect, higher behavioral dimensions are present in SNNs compared to non-spiking \gls{dnns}. Therefore, a more comprehensive study is required to understand the inherent behavior of \gls{snns}, especially under adversarial attacks. Towards this, the following key questions need to be investigated:\\ 
\noindent \textbf{(Q1)} \textit{How do the spiking structural parameters (i.e., threshold voltage and time window\footnote{The \textit{threshold voltage} is defined such that, when the spiking neuron's membrane potential overcomes such threshold, the neuron emits an output spike and resets its membrane potential. The \textit{time window} represents the observation period in which the SNN receives the same input.}) affect the \gls{snns}' behavior under attack?}\\
\noindent \textbf{(Q2)} \textit{Are \gls{snns} inherently robust against adversarial attacks, regardless of the structural parameters?}\\
\noindent \textbf{(Q3)} \textit{Does a combination of structural parameters that provides high accuracy also guarantee high robustness?}

%Another key research challenge that we want to study in this work is the impact of the structural parameters of the \gls{snns} on its security. ``Can the inherent structural parameters of the \gls{snns} be adjusted towards achieving higher adversarial robustness w.r.t. the \gls{snns} with the default settings?''

\subsection{Motivational Case Study}

\begin{figure}[!htp]
	\centering
	\includegraphics[width=\linewidth]{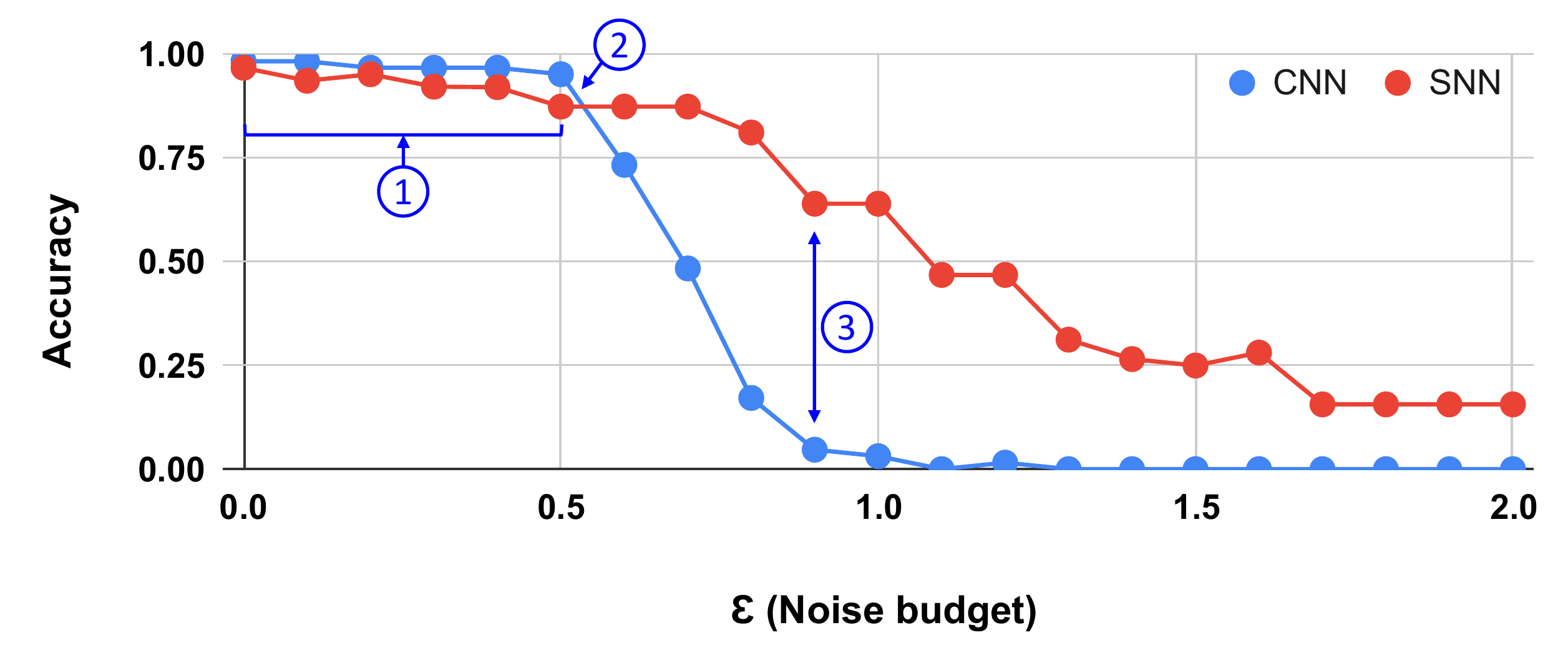}
	\vspace*{-20pt}
	\caption{PGD adversarial attack applied to a \gls{cnn} and an SNN that have the same number of layers with equal size and equal number of neurons.}
	\label{fig:CNN_SNN_PGD2} 
\end{figure}

Before answering the above-discussed questions, we conducted a motivational case-study to highlight the importance of the problem. We trained a 5-layer \gls{cnn}, with 3 convolutional layers and 2 fully-connected layers on the MNIST dataset~\cite{mnist} using the PyTorch framework~\cite{PyTorch}, and an SNN with the same number of layers and neurons per layer with the Norse framework~\cite{norse}. We applied the white-box PGD attack~\cite{pgd} on both networks and monitored the accuracy variation w.r.t. the noise budget $\varepsilon$. The results reported in Figure~\ref{fig:CNN_SNN_PGD2} indicate that, while with low noise magnitude the CNN has higher accuracy (pointer~\rpoint{1}), after the turnaround point of $\varepsilon = 0.5$ (pointer~\rpoint{2}), the SNN clearly shows a more robust response to the attack than its \gls{cnn} counterpart, with an accuracy gap higher than 50\% (pointer~\rpoint{3}). For $\varepsilon > 0.5$, the accuracy of the \gls{cnn} decreases sharply, while the slope for the SNN is lower. This outcome motivated us to further investigate the inherent robustness of the \gls{snns}.

These experiments give a quick overview on the high potential of \gls{snns} in terms of security when compared to traditional \gls{dnns}. However, while these experiments are run using the default SNN structural parameters, one cannot generalize this observation until a deeper analysis is made. 

%(If possible, insert a result figure in this section.) \newline
% \alberto{@Rida: could you please add a figure with the response of PGD or FGSM attack of an SNN and a DNN?}\newline
% \rida{@Alberto: The figure is the robustness on PGD.  The combination for the SNN here is $T=64;v_th=1.5$ , do you want me to add it on the figure?}\newline
% \alberto{@Rida: thank you, that's what we need in the motivational analysis.}
\subsection{Our Novel Contributions}
%In this paper, we investigate the \gls{snns} robustness to adversarial attack and present the following contributions, which are summarized in Figure~\ref{fig:novel_contributions}.

\begin{figure}[b]
	\centering
	\includegraphics[width=.95\linewidth]{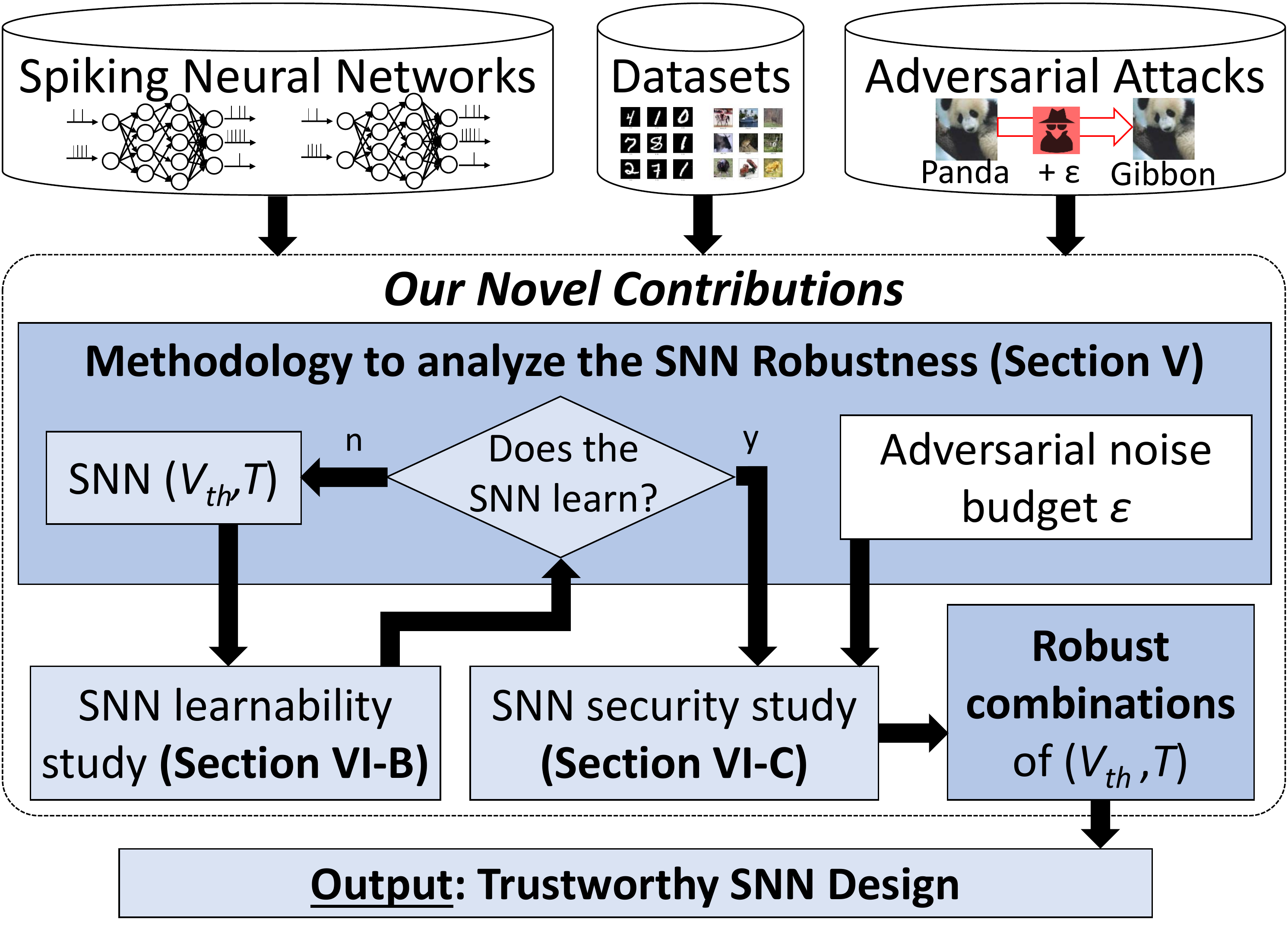}
	\vspace*{-8pt}
	\caption{Overview of our novel contributions (shown in blue boxes).}% \ihsen{can we improve the figure? @Alberto, can you share the figure source please, or work on it if you want? thanks} \alberto{I shared all the sources in the shared folder on Drive, but if you want I can modify this figure once the methodology diagram is done. I think the methodology block here should reflect the detailed diagram shown in the later section.}}
	\label{fig:novel_contributions}
\end{figure}

\begin{itemize}[leftmargin=*]

    \item We propose a systematic methodology (see Figure~\ref{fig:novel_contributions}) for analyzing the SNN robustness. We are first to explore the impact of neurons' structural parameters (i.e., spiking threshold voltage $V_{th}$ and time window boundary $T$) on the \gls{snns}' robustness against the strong white-box adversarial attacks. \textbf{[Section~\ref{sec:methodology}]}
    \item The SNN learnability and security studies show that the \gls{snns}' inherent robustness is strongly conditioned by these structural parameters. \textbf{[Section~\ref{sec:experiments}]} %Importantly, sweet spots of the combination of threshold voltage and time window are found. 
    \item We design trustworthy \gls{snns}, by fine-tuning their structural parameters around the previously-found sweet spots. For instance, a 5-layer SNN trained on MNIST dataset achieves up to $84\%$ accuracy improvement compared to a corresponding CNN, when the PGD attack with $\varepsilon=1.5$ noise budget is applied. \textbf{[Section~\ref{subsec:security}]} %\alberto{@Rida: could you please replace the TODOs with an example taken from the results?}
    \item \textbf{Open-Source Contribution:} for reproducible research, we release the complete source code of our methodology, including all the examples and robust SNN models, at \url{https://github.com/rda-ela/SNN-Adversarial-Attacks}. %\alberto{@Ihsen and @Rida: Should we keep the open-source claim or not?} \ihsen{I don't mind keeping it.}
    %\item We test the robustness of DNNs and SNNs, by applying white-box adversarial attacks on spiking and non-spiking networks with the same layer and dimension structures, towards a fair comparison.
    %\item We perform a DNN-to-SNN transferability analysis w.r.t. the robustness against white-box adversarial examples.
\end{itemize}

\section{Background }
\subsection{Spiking Neural Networks}

\gls{snns} are considered as the third generation of neural networks~\cite{Maas1997ThirdGenerationSNN}. Their event-based communication scheme between different neurons is inspired by the human brain's functionality with a higher level of similarity than the (non-spiking) \gls{dnns}. Such a bio-inspired computation not only provides biologically-plausible deep learning, but also represents a huge potential for bridging the energy-efficiency gap between the human brain and the supercomputers executing complex deep learning applications, as motivated by the recent advances of neuromorphic architectures, such as IBM TrueNorth\cite{Merolla2014Truenorth} and Intel Loihi\cite{Davies2018Loihi}.

\begin{figure}[h]
	\centering
	\includegraphics[width=\linewidth]{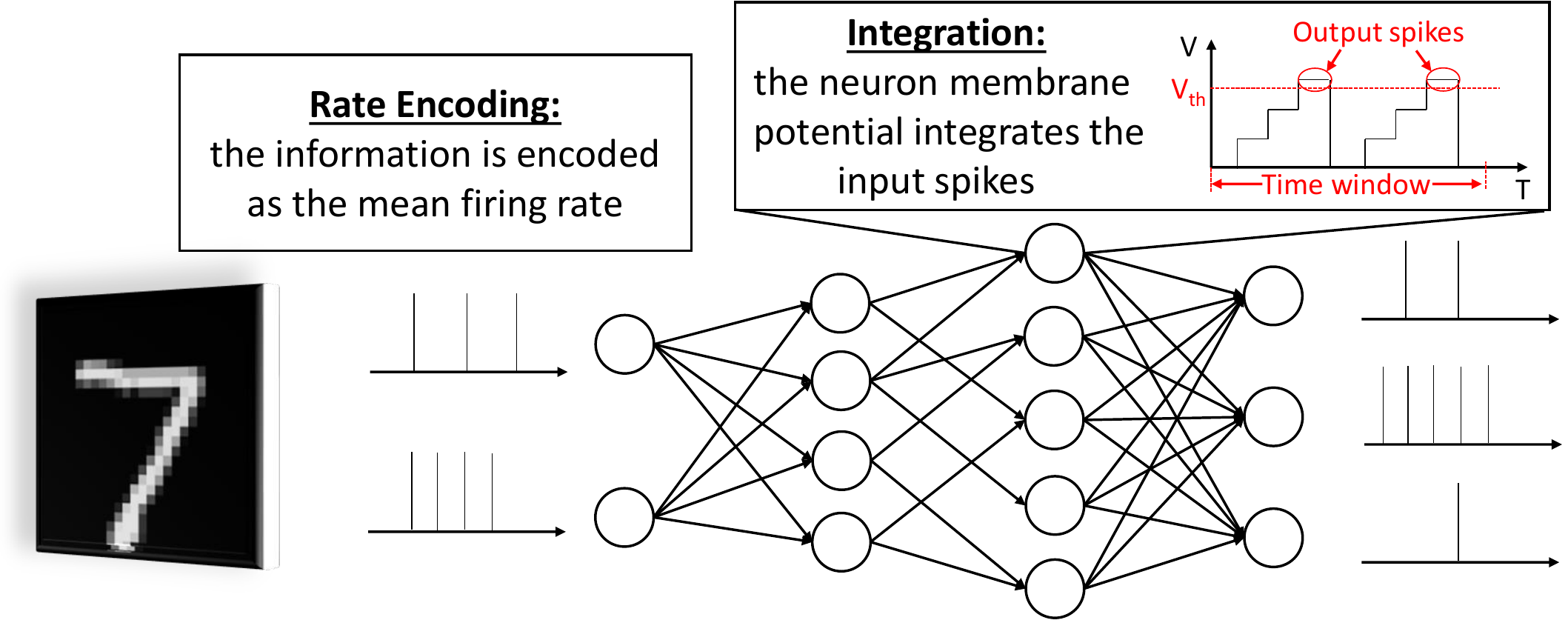}
	\vspace*{-15pt}
	\caption{Overview of the functionality of an SNN, with a focus on the rate encoding of the information and the integration of the spikes into the membrane potential.}
	\label{fig:SNN}
\end{figure}

An example of the \gls{snns}' modus operandi is illustrated in Figure~\ref{fig:SNN}. The input information has to be properly coded using spikes. While other possible coding schemes can be based on the delay between consecutive spikes or the latency between the beginning of the stimulus to the first spike, the most commonly adopted mechanism is the \textit{rate encoding}~\cite{Kasinski2011IntroSNN}, where the activation intensity corresponds to the mean firing rate over a certain \textit{time window}. Such a time window represents the observation period in which the SNN receives the same input. A wider window gives more time for the spikes to propagate towards the output, but incurs in higher latency. When an incoming spike $s_i$ arrives at the input of the neuron, it is multiplied by its associated synaptic weight $w_i$ and integrated into the membrane potential $V$, following Equation~\ref{eq:neuron}.

\begin{equation}
\label{eq:neuron}
V = \sum_{i=1}^N w_{i} \cdot s_{i}
\end{equation}

For a Leaky-Integrate-and-Fire (LIF) spiking neuron model, when the membrane potential exceeds the \textit{threshold voltage} $V_{th}$, the neuron emits an output spike and resets its membrane potential. In this way, the information is propagated to the output. For the example of rate encoding for image classification, the higher the firing rate of the output spike train is, the higher the output probability of the associated class.

Moreover, supervised learning for \gls{snns} is complex, because of the non-differentiability of the SNN loss function~\cite{Bodo2019ClosingTheAccuracyGap}. Therefore, the standard backpropagation procedure cannot be applied. To overcome these challenges, there are two possibilities. The first solution consists of training a correspondent DNN with standard backpropagation, and then converting it to the spiking version~\cite{Massa2020EfficientSNN}. However, a certain accuracy is typically lost in the conversion process. Another possibility is to approximate the SNN derivatives and learn based on the temporal information in the spiking domain~\cite{Thiele2020SpikeGrad}. \textit{The latter option is adopted in our work.}

%\ihsen{@Alberto: Could you please take care of this subsection? We need to explain the neuron's firing process, the community is not necessarily expert yet in \gls{snns}. It needs to be understandable.} \alberto{ok}
\subsection{Adversarial Attacks}
\gls{dnns} are now deployed in a wide range of sectors, including safety-critical applications such as Intelligent Transportation Systems~\cite{signs}. Despite their performance, \gls{dnns} suffer from a critical challenge, i.e.,  adversarial attacks. Many studies~\cite{vulnerable, pgd, neuroattack, RobustML2_shafique, Khalid2019FAdeML, Marchisio2019CapsAttacks} have shown that \gls{dnns} are vulnerable to carefully crafted inputs designed to fool them, very small imperceptible perturbations added to the data can completely change the output of the model.

\noindent 
{\bf Minimizing injected noise:} It is essential for an attacker to minimize the added adversarial noise to avoid detection.  
%, an adversary, using information learned about the structure of the classifier, tries to craft perturbations added to the input to cause incorrect classification.   In these types of attacks, the adversary desires to minimize this perturbation noise to avoid detection.  
Formally, given an original input $x$ with a target label $l$ with a classification model $ f() $, the problem of generating an adversarial example $x^*$ can be formulated as a constrained optimization problem~\cite{pbform}, i.e., 

\begin{equation}
\label{eq:adv}
     \begin{array}{rlclcl}
        x^* = \displaystyle \argmin_{x^*} & \mathcal{D}(x,x^*),  \\
         s.t.~  & f(x) = l,  ~ f(x^*) = l^*,  ~ l \neq l^*
\end{array}
\end{equation}

Where $\mathcal{D}$ is the distance between two images and the optimization objective is to minimize this adversarial noise to make it stealthy. $x^*$ is considered as an adversarial example if and only if $ f(x) \neq  f(x^*) $ and the noise is bounded ($\mathcal{D}(x,x^*) < \epsilon $, where $\epsilon \geqslant 0 $).

%\subsection{Distance metrics}
% \noindent
% {\bf Distance Metrics:}
% The adversarial examples and the added perturbations should be visually imperceptible by humans. Three main metrics are generally used to approximate the visual difference, namely $L_0$, $L_2$, and $L_ \infty$~\cite{carlini2016evaluating}. These metrics are special cases of the $L_p$ norm: 

% \begin{equation}
%     \left\|x\right\|_p = \left( \sum^{n}_{i = 1} \left |x_i \right | ^{p} \right)^{\frac{1}{p}}
% \end{equation}

%These metrics focus on different aspects of visual significance. $L_0$ counts the number of pixels with different values at corresponding positions in the two images. $L_2$ measures the Euclidean distance between the two images $x$ and $x^*$. $L_ \infty$ measures the maximum difference for all pixels at corresponding positions in the two images.

%Still in the financial sector, companies frequently use graph convolutional networks (GCNs)~\cite{kipf} to decide whether their customers are trustworthy or not. If scammers succeed in hiding their personal identity information, and avoid company's detection, this will result in a significant loss to the company. Therefore, the security issues of deep neural networks represents a major concern.

\section{Related Work}

Recent studies analyzed the \gls{snns}' robustness. 
Schuman et al.~\cite{Schuman2020ResilienceAR} analyzed the resilience of \gls{snns} to faults, varying the training method. 
In NeuroAttack~\cite{neuroattack}, the impact of the bit-flips on the accuracy of \gls{snns} was studied. The adversarial perturbation was introduced to fire a target neuron to trigger a hardware Trojan.

Adversarial attacks have been tested on \gls{snns}. Bagheri et al.~\cite{Bagheri2018AdvTrainingSNN} studied the sensitivity of SNN w.r.t. different types of encoding, when subjected to white-box adversarial attacks. 
Marchisio et al.~\cite{Marchisio2019SNNAttack} applied black-box adversarial attacks to \gls{dnns} and \gls{snns}, and the comparison showed that the \gls{snns} are more robust. 
Sharmin et al.~\cite{Sharmin2019ACA} proposed a methodology to perform the adversarial attack on (non-spiking) \gls{dnns}, and then the DNN-to-SNN conversion made the adversarial examples craft the \gls{snns}. 
Liang et al.~\cite{Liang2020ExploringAA} proposed a gradient-based adversarial attack methodology for \gls{snns}, and showcased the impact of the adversarial attack success rate on the type of loss function and threshold voltage of the second-last layer only. 
The work of~\cite{Sharmin2020InherentAR} analyzed the adversarial accuracy of \gls{snns} trained with different inference latency and leak factor in LIF spiking neurons. However, this work did not explore the impact of membrane voltage threshold along with the time window. \textit{Note, in contrast to the work in~\cite{Sharmin2020InherentAR}, we explore the impact of spiking parameters of the neurons on \gls{snns} robustness to question the generalization of inherent robustness observation}. 

Recently, Massa et al.~\cite{Massa2020EfficientSNN} tuned the threshold voltage and the time window with the purpose of minimizing the accuracy loss in the DNN-to-SNN conversion process. This work did not study robustness to adversarial attacks.
DIET-SNN~\cite{rathi2020dietsnn} proposed to tune the membrane threshold and membrane leak to jointly optimize the accuracy and the latency. \textit{However, none of these prior works analyzed the impact of structural parameters, i.e., membrane threshold and time window, from the security perspective, as we target in this paper.}

\section{Threat Model}
\subsection{Adversary Knowledge}

\begin{figure}[htp]
	\centering
	\includegraphics[width=\linewidth]{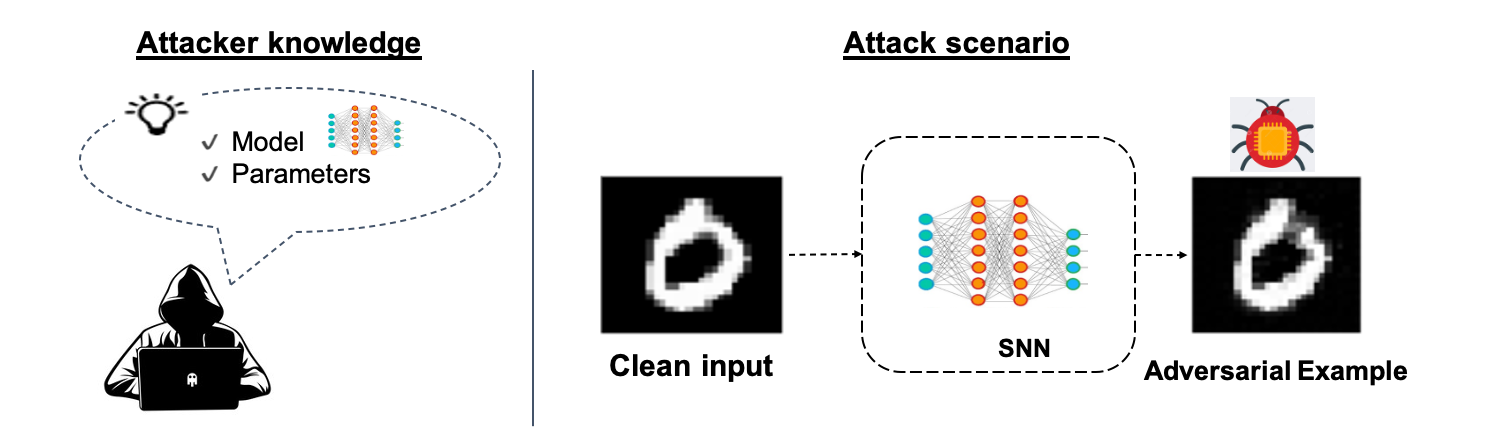}
	\vspace*{-16pt}
	\caption{White-box setting: The attacker has full access to the SNN structure, input \& output, hyper parameters, weights, and even structural parameters.}
	\label{fig:whitebox}
\end{figure}

In our experiments, we assume the strongest case where an attacker is attempting to design adversarial attacks to fool a SNN classifier in a white-box attack scenario. In fact, we assume a powerful attacker who has the full knowledge of the victim classifier's architecture and parameters (including the structural parameters $V_{th}$ and $T$).  The attacker uses this knowledge to create adversarial examples. Figure~\ref{fig:whitebox} gives an overview of the attack scenario.

\subsection{Attack Generation}

%\ihsen{we might probably delete FGSM if we don't report results of it later.}
%\noindent \textbf{Fast Gradient Sign Method (FGSM).} The Fast Gradient Sign Method~\cite{fgsm} is a single-step, gradient-based, attack. An adversarial example is generated by performing a one step gradient update along the direction of the sign of gradient at each pixel as follows:

%  \begin{equation}
%      x^* = x + \epsilon \cdot sign (\nabla_{x}\mathcal{L}_{\theta}(x,y))
%  \end{equation}
% Where $\nabla \mathcal{L}()$ computes the gradient of the loss function $\mathcal{L}$ and $\theta$ is the set of model parameters. The $sign()$ denotes the sign function and $\epsilon$ is the perturbation magnitude. 
We evaluate the \gls{snns} robustness using one of the most widely used attacks, namely PGD~\cite{pgd}. It is one of the strongest iterative variant of the FGSM where the adversarial example is generated as follows:
 \begin{equation}
x^{t+1} = \mathcal{P}_{\mathcal{S}_x}(x^t + \alpha \cdot sign (\nabla_{x}\mathcal{L}_{\theta}(x^t,y)) )
 \end{equation}
Where $\mathcal{P}_{\mathcal{S}_x}()$ is a projection operator projecting the input into the feasible region $\mathcal{S}_x$ and $\alpha$ is the additive noise at each iteration.
PGD attack tries to find the perturbation that maximizes the loss of a model on a given sample while keeping the perturbation magnitude lower than a given budget. PGD is an iterative gradient-based attack that is considered as a high-strength attack~\cite{RobustML_shafique, RobustML2_shafique}.

%Table~\ref{Attack_methods} gives an overview of key characteristics of PGD attack. 

% \begin{table}[!htp]
% \centering
%   \caption{Summary of the used attack methods. (the strength estimation is based on~\cite{strength})} 
%   \label{Attack_methods}
%   \begin{tabular}{ccccc}
%     \toprule
%     \textbf{Method} & \textbf{Category}  & \textbf{Perturb. Norm} & \textbf{Learning} & \textbf{Strength} \\
%     \midrule
%     PGD   & gradient-based   &  $L_\infty$  & Iterative & high \\
% %    Universal perturbations & & Untargeted & Universal & $L_2$ & Iterative & *****\\
    
%   \bottomrule
% \end{tabular}
% \end{table}

\section{Proposed Methodology} \label{sec:methodology}
\begin{figure}[!h]
	\centering
	\includegraphics[width=0.9\linewidth]{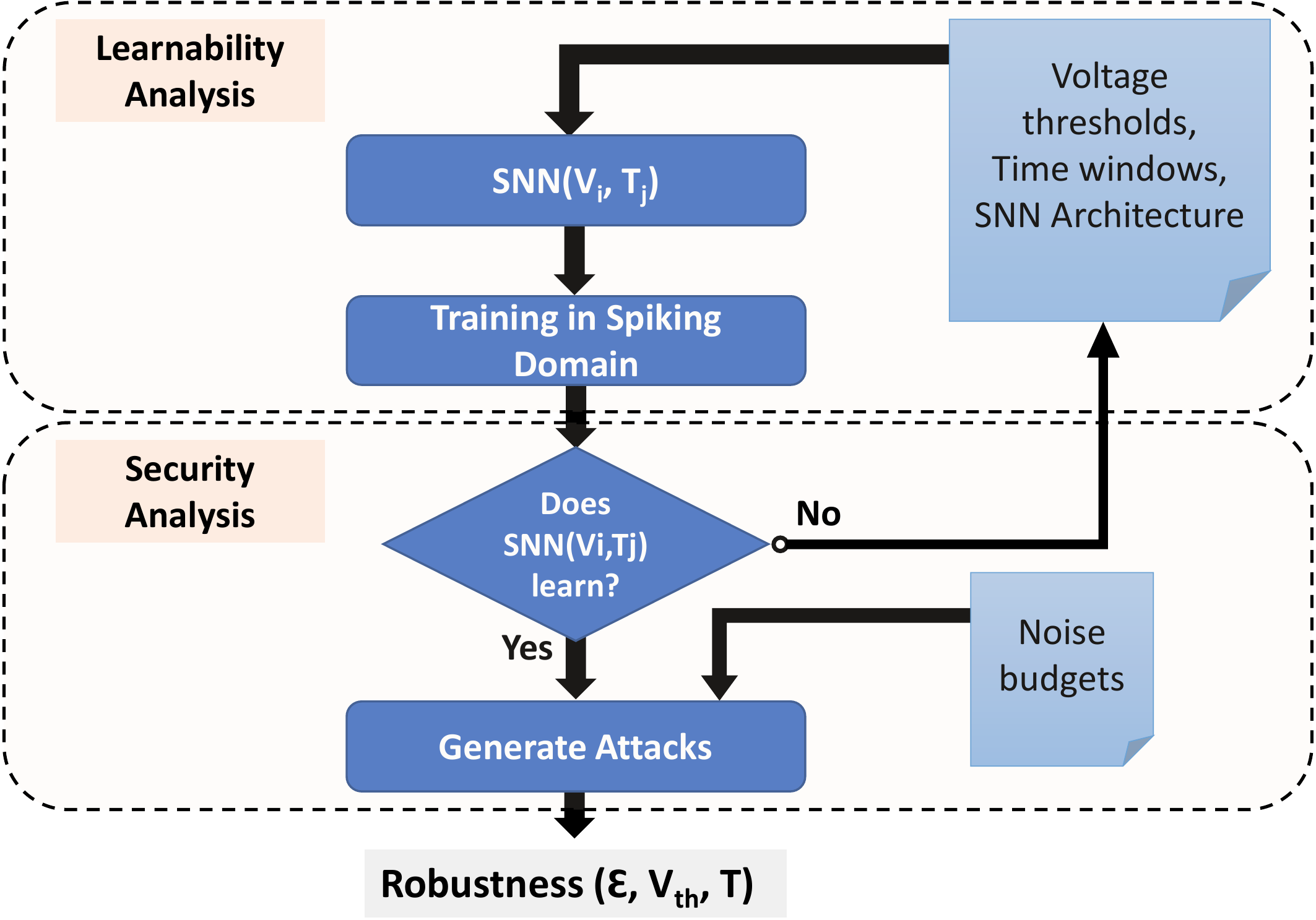}
	\vspace*{-6pt}
	\caption{A flowchart illustrating the key steps of SNN robustness exploration.} 
	\label{fig:flow}
\end{figure}

Our study aims at the exploration of \gls{snns}' robustness under different adversarial noise budgets, and this, for different $(V_{th}, T)$ parameters combinations. Figure \ref{fig:flow} gives an overview of different components of our methodology. It is composed of the following two main steps. \textbf{(1)} The first step of the exploration is meant to exclude combinations that are not propitious for efficient learning in SNNs. There is indeed no interest in studying the robustness of SNNs with low baseline performance. \textbf{(2)} In the second step, for all $(V_{th}, T)$ settings that enable the SNN training to converge efficiently, we proceed to a robustness exploration.

Algorithm \ref{algo:robust} details our robustness exploration methodology. Line $1$ and $2$ browse the $n$ threshold voltages and $m$ time windows to explore. Once the training is launched (Line $3$), we proceed to the SNN learnability study for the given combination ($V_{th},T$). As shown in Line 4, the learnability is quantitatively verified by setting a minimum baseline accuracy level below which we consider the SNN learning inefficient. This value depends on the given SNN architecture, its learning method, the dataset and the application. In our case study, we use this accuracy threshold equal to $70\%$ as it is typically achieved by state-of-the-art SNNs. The security analysis starts from Line $5$; it consists of generating adversarial examples with different noise budgets to fool the SNN. The noise budget models the aggressiveness allowed within the attack generation; the higher the noise budget, the more aggressive the attack is considered. First, the counter of successful attack generation cases is initialized (Line 6). Then, we browse the dataset $\mathcal{D}$ (Line 7) to generate the adversarial attacks using PGD, as shown in Line $8$. Afterwards, the algorithm verifies if the generated example is able to fool the SNN (Lines 9 - 13), i.e., if the attack succeeded to force the output to a wrong label, and accordingly increment the adversarial success counter. Then, the robustness is then evaluated for every $\varepsilon$ value as the rate of attacks for which the adversary failed to generate an adversarial example that fools the victim SNN (Line 15). Hence, by tracking the accuracy slope with regard to $\varepsilon$, we can compare the robustness of each model to adversarial attacks. 

% \alberto{@Ihsen: I think we need a schematic diagram here, because otherwise the methodology section is too weak. Can you please take care of it?} \ihsen{ok}

\begin{algorithm}[!htp]
\SetAlgoLined
\begin{small}
\KwResult{Robustness Level }
\KwData{\\ Membrane Voltage Thresholds: $V{th} = V_{i} ~/ i \in [1,n] $; \\ Spiking Time Windows: $ T= T_{j}/ j \in [1,m]$; \\ Adversarial Noise Budgets: $\varepsilon = \varepsilon_k ~/ k \in [1,p]$;\\ SNN Architectures: $S_{ij} = SNN(V_i, T_j)$; \\ Labeled Test Set: $\mathcal{D} = (X_t, L_t)$ ;\\ Accuracy threshold: $A_{th}$ }

% initialization\;
 \For{$i \leftarrow 1$ \KwTo $n$}{
     \For{$j \leftarrow 1$ \KwTo $m$}{
        Train($S_{ij}= SNN(V_i, T_j)$)\;
        \eIf{Accuracy($S_{ij}) \geq A_{th}$}{ 
            \tcp{$S_{ij}$ learns}
            \For{$k \leftarrow 1$ \KwTo $p$}{
                Adv = 0\;
                \For{$X_t \leftarrow 1$ \KwTo $<\mathcal{D}>$}{
                    \tcp{Adversarial Attack}
                    $X^*_t = PGD(S_{ij}, \varepsilon_k, X_t)$\;
                    \eIf{$S_{ij}(X^*_t) \neq L_t$}{
                        Adv++ \;
                        }{NOP\;}
                }
            Robustness ($\varepsilon_k$)= $1- \frac{Adv}{<\mathcal{D}>}$\;
            }
            
        }{NOP\;}
    }

 }
 \end{small}
 \caption{Robustness Exploration Algorithm.}\label{algo:robust}
\end{algorithm}

%\subsection{White-Box Attack Generation}
%\subsection{DNN-to-SNN Transferability}
%\subsection{Robustness w.r.t. the SNN structural parameters}

\section{Evaluation}\label{sec:experiments}

\subsection{Experimental Setup}
Our experiments are performed using Norse \cite{norse} which is a library that expands PyTorch \cite{PyTorch} with primitives for bio-inspired neural components, thereby allowing to train and run SNNs in the spiking domain. The adversarial attacks are implemented using Foolbox v3.1.1 \cite{foolbox}. The SNN architecture is a Lenet-5 adapted to the spiking domain, and trained on the MNIST database \cite{mnist}. Note: unlike \gls{dnns}, SNNs are still in a relatively early phase of adoption. We adopt the same test conditions as widely used by the SNN research community where the typical evaluation settings~\cite{Allred2020ControlledForgetting} use datasets like MNIST and Fashion MNIST. The used neuron model is the Leaky-Integrate-and-Fire (LIF) Neuron and the experiments were run on an Nvidia TESLA P100 GPU with a 16GB memory.

\subsection{Learnability Study}
\begin{figure}[ht]
	\centering
	\vspace*{-8pt}
	\includegraphics[width=\linewidth]{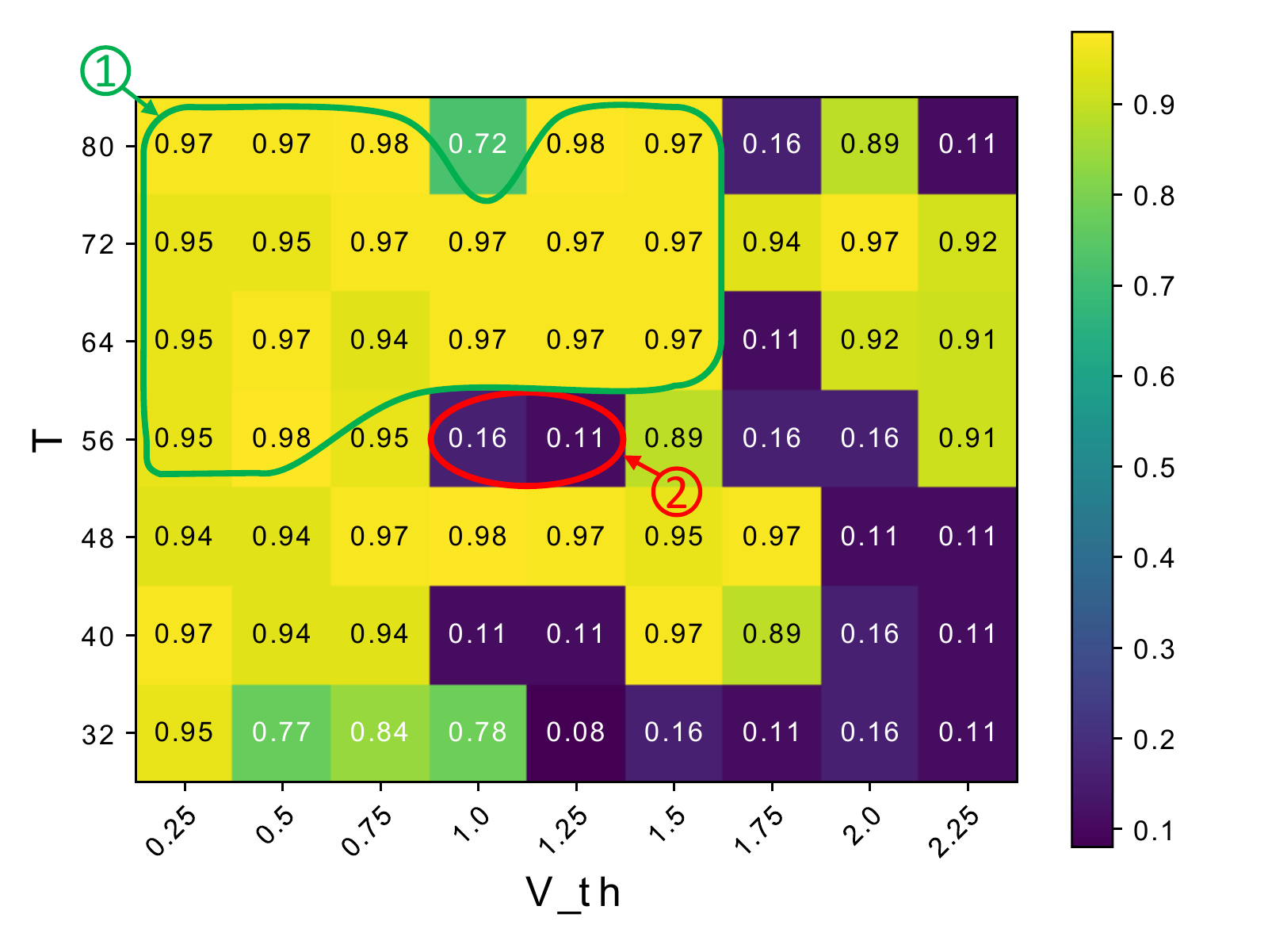}
	\vspace*{-24pt}
	\caption{A heat map showing the accuracy of \gls{snns} trained on MNIST dataset under different combinations of $V_{th}$ and $T$.} %\ihsen{@Rida, could you please export the figure from python in pdf? and could you please add axes legend? thanks}}
	\label{fig:acc_heat}
	\vspace*{-8pt}
\end{figure}

%Here comes the heat map experiments, as a first exploration step 
Before studying the robustness with respect to varying the structural parameters, we need to define our exploration space. In fact, the default values of the threshold voltage and time window parameters are $(V_{th} , T) = (1, 64)$. Therefore, we focus on having an overview of the learnability of \gls{snns} in the neighborhood of this setting. Figure~\ref{fig:acc_heat} shows the accuracy heat map for different  $(V_{th} , T)$ combinations. The horizontal and vertical axes denote $V_{th}$ and $T$, respectively. Different colors denote the accuracy of the SNN. 
Note from pointer~\rpoint{1} of Figure~\ref{fig:acc_heat} that the highest-accuracy combination tends to be towards the top-left corner, i.e., low $V_{th}$ and high $T$. However, the heat map is clearly not monotonic. For example, as indicated by pointer~\rpoint{2}, there are combinations with an accuracy lower than $16\%$ which are surrounded by combinations with accuracy higher than $89\%$.

While it is obvious that studying robustness for the non-learnable combination is not useful, we use this map as a reference to track the behavior of \gls{snns} under attack with different noise budgets. 

%We might mention sweet spots, but also bad surprises as limitation of previous work;

\subsection{Security Study} \label{subsec:security}

\begin{figure}[ht]
	\centering
	\vspace*{-8pt}
	\includegraphics[width=\linewidth]{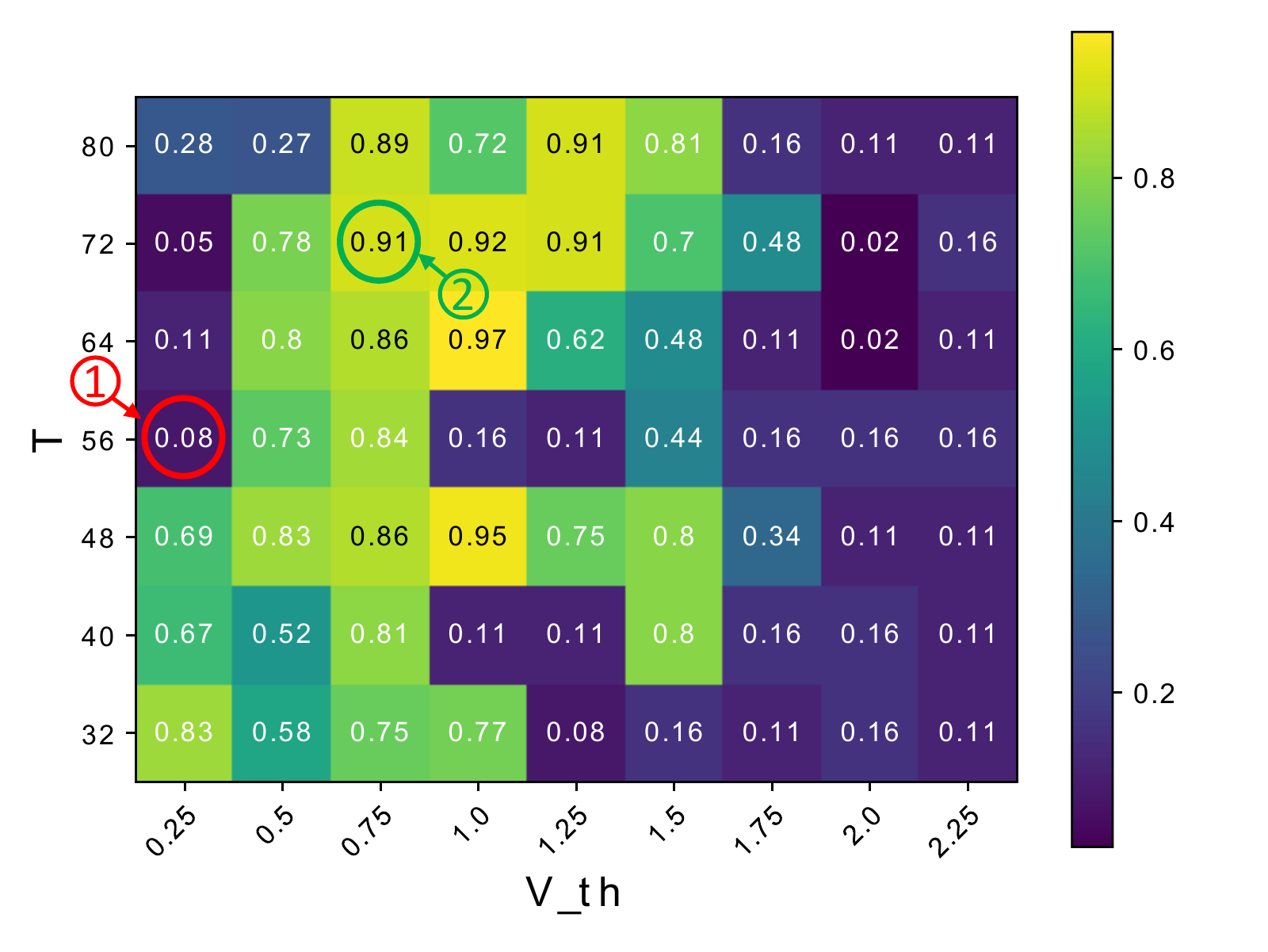}
	\vspace*{-24pt}
	\caption{A heat map showing the accuracy of \gls{snns} trained on MNIST datatset under different combinations of $V_{th}$ and $T$ under PGD attack with \underline{$\varepsilon = 1$}. }%\ihsen{@Rida, could you please export the figure from python in pdf? and could you please add axes legend? thanks}}
	\label{fig:eps1_heat}
\end{figure}

\begin{figure}[h]
	\centering
	\includegraphics[width=\linewidth]{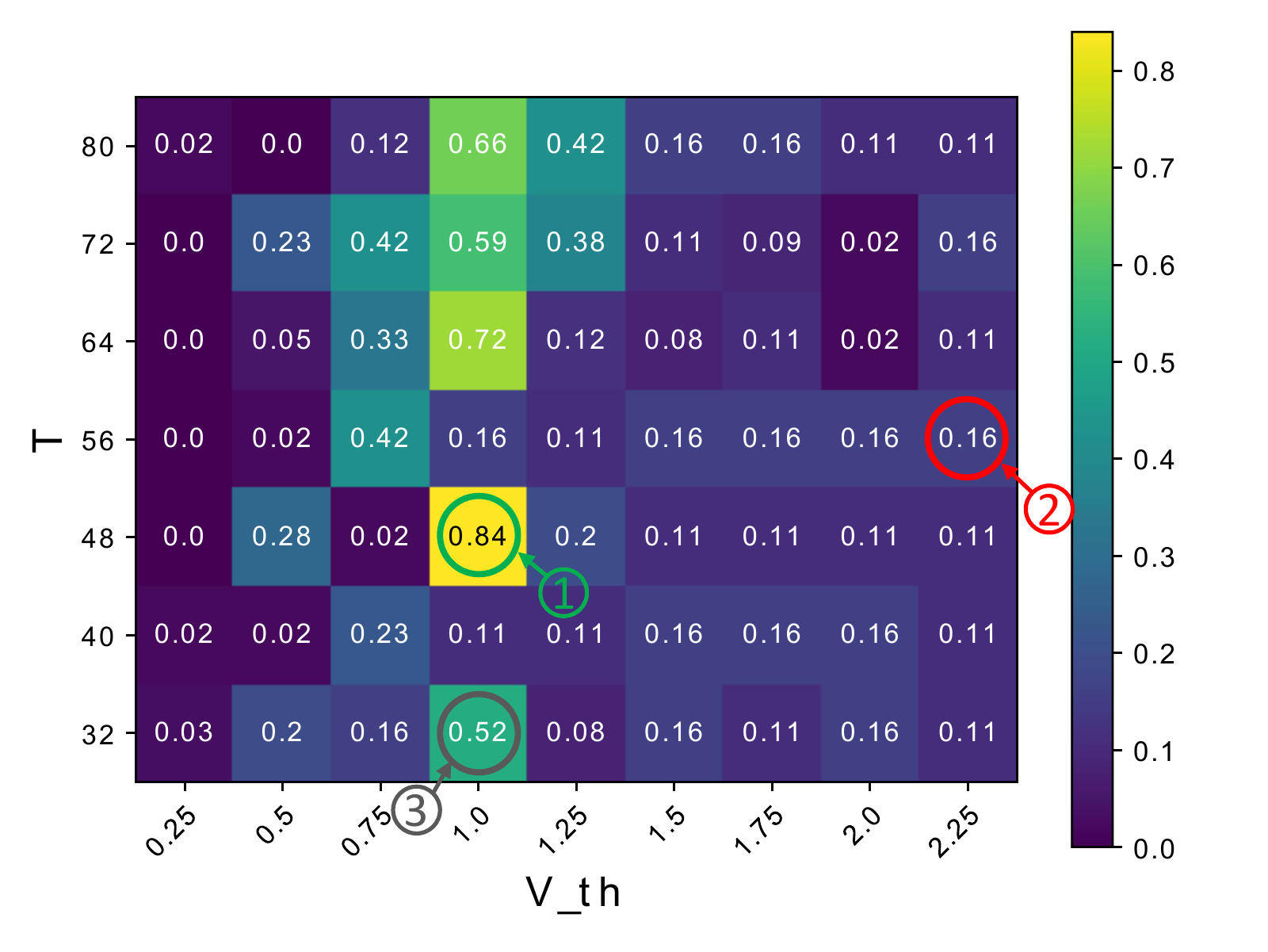}
	\vspace*{-20pt}
	\caption{A heat map showing the accuracy of \gls{snns} trained on MNIST dataset under different combinations of $V_{th}$ and $T$ under PGD attack with \underline{$\varepsilon = 1.5$}. }%\ihsen{@Rida, could you please export the figure from python in pdf? and could you please add axes legend? thanks}}
	\label{fig:eps15_heat}
\end{figure}

In this section, we investigate the robustness of \gls{snns} while increasing the attacks' adversarial noise magnitude in a white-box scenario. We first proceed to a holistic exploration under all previous combinations of $V_{th}$ and $T$. Figures~\ref{fig:eps1_heat} and~\ref{fig:eps15_heat} show the accuracy degradation of \gls{snns} under PGD attack with noise magnitudes of $1$ and $1.5$, respectively.

The first interesting insight that we extract from Figures~\ref{fig:eps1_heat} and~\ref{fig:eps15_heat} is that, high baseline learnability (without adversarial attacks) is not a guarantee of robustness. Moreover, we notice a different evolution of the \gls{snns} w.r.t. adversarial attacks based on their respective structural parameters.
More specifically, two \gls{snns} with a starting comparable accuracy may have different behaviors under attack. For example, combinations $(V_{th},T) = (0.25,56)$ and $(V_{th},T) = (0.75,72)$ start with respective accuracy of $95\%$ and $97\%$. However, while the accuracy of the first combination (pointer~\rpoint{1} of Figure~\ref{fig:eps1_heat}) drops drastically to $8\%$ under $\varepsilon=1$ attack budget, the second (pointer~\rpoint{2}) looses only $6\%$ of its initial accuracy under the same attack noise magnitude. 
Moreover, from Figure~8 we can observe different types of robustness to the attack:
\begin{enumerate}
    \item $(V_{th},T) = (1,48)$: high robustness
    \item $(V_{th},T) = (2.25,56)$: low robustness
    \item $(V_{th},T) = (1,32)$: medium robustness
\end{enumerate}

\begin{figure}[h]
	\centering
	\vspace*{-8pt}
	\includegraphics[width=\linewidth]{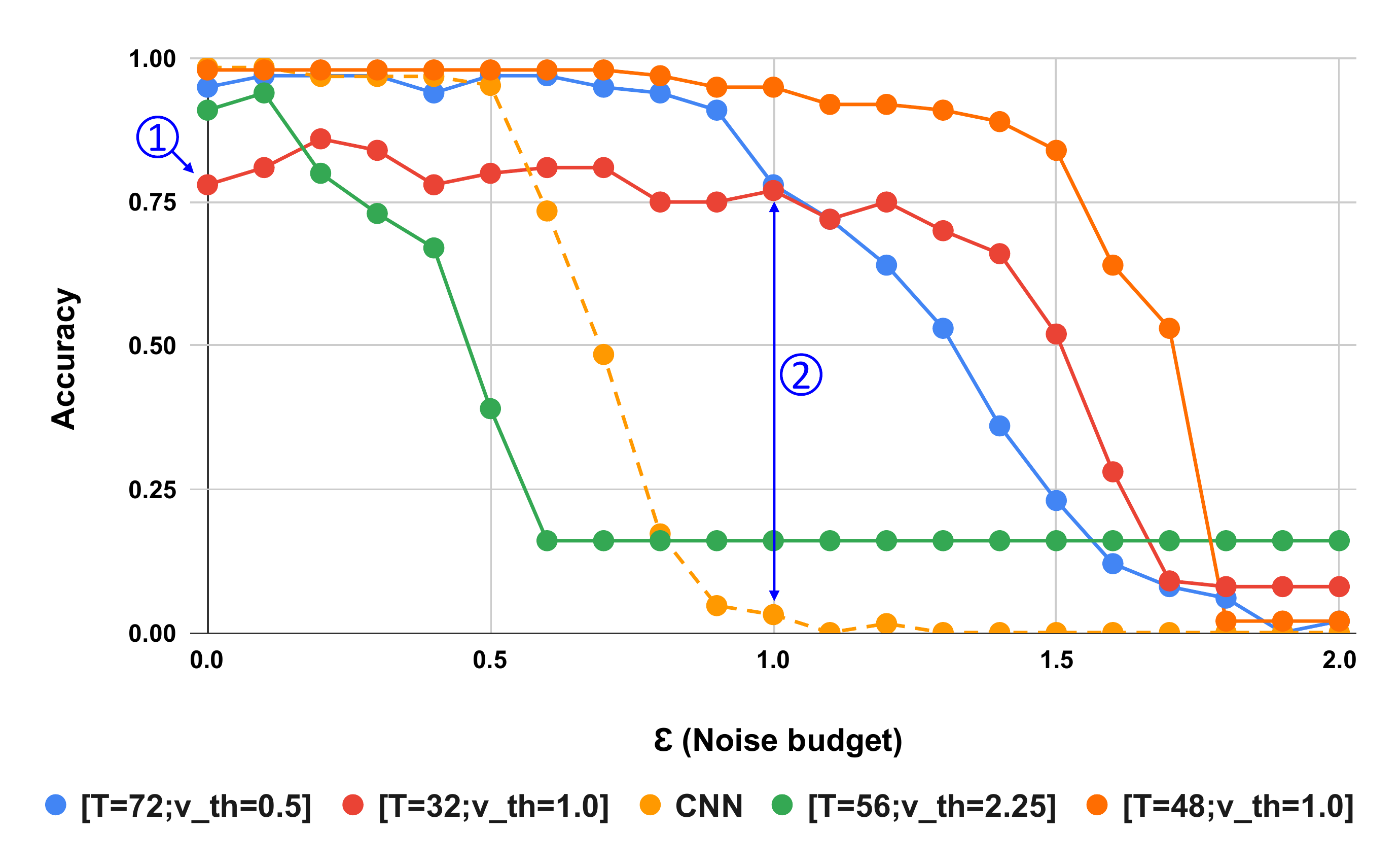}
	\vspace*{-20pt}
	\caption{Robustness of \gls{snns} tested on MNIST with different $V_{th}$ and $T$ parameters under PGD attack compared to the Lenet-5 \gls{cnn}.}%\ihsen{@Rida: the combination (1,48) looks from the heat maps the best, could you please add it to the figure (even replacing (1,72) if the figure becomes unclear)}\newline	\rida{@Ihsen: done}}
	\label{fig:epsilon1} 
\end{figure}

In the following, we track a set of insightful $(V_{th} , T)$ combinations and track their impact on \gls{snns} robustness compared to the Lenet-5 \gls{cnn} trained on the same dataset. Figure~\ref{fig:epsilon1} compares the robustness of \gls{snns} with different structural parameters w.r.t. its correspondent \gls{cnn}. This figure shows, in a more detailed fashion, the impact of structural parameters on \gls{snns}' security. In fact, while the combination $(V_{th}, T) = (2.25, 56)$ achieves lower robustness than the CNN, up to $85\%$ higher robustness is reached by the combination $(V_{th},T)=(1,48)$. Another interesting case is represented by the combination $(V_{th},T)=(1,32)$, whose clean accuracy is only $78\%$ (see pointer~\rpoint{1} in Figure~\ref{fig:epsilon1}), while, as indicated by pointer~\rpoint{2}, it has $75\%$ higher accuracy than the CNN when a strong noise budget (i.e., $\varepsilon>1$) is applied.

\section{Concluding Remarks} \label{sec:disc}
This paper investigated the security of \gls{snns} from a new perspective., i.e., a systematic exploration of the impact of structural parameters of the bio-inspired neurons on the robustness of \gls{snns} to adversarial attacks. Under a strong attack scenario, i.e., white-box setting, we performed extensive security analysis with respect to three variant parameters: spiking voltage threshold ($V_{th}$), spiking time window ($T$) and attack's noise budget ($\varepsilon$). By tracking the robustness considering those parameters, we found a high impact of $V_{th}$ and $T$ on the robustness of \gls{snns}. While this work confirms state-of-the-art claims about the inherent \gls{snns} robustness, it generalizes those findings and shows their relativity to spiking structural parameters, which is the first study revealing these valuable insights. 
In summary, this work answers the three questions raised in the introduction (Section \ref{sec:intro}-A) as follows:\\
\noindent \textbf{(A1)} \textit{Structural parameters ($V_{th}, T$) do have a significant impact on the robustness of \gls{snns}}, and a careful exploration needs to be carried out before deployment in safety-critical or security-sensitive applications.\\ 
\noindent \textbf{(A2)} Yes, SNNs have inherent robustness against adversarial attacks. \emph{However, this inherent robustness is highly conditioned by the choice of ($V_{th}, T$) combination.}\\
\noindent \textbf{(A3)} No, a combination of ($V_{th}, T$) parameters that learns efficiently and gives high baseline accuracy is \textit{not} a guarantee of robustness. 

These findings consolidate the positioning of SNNs as an interesting solution towards efficient and robust  machine learning systems. SNNs' high power efficiency makes them even more interesting, especially for embedded systems and at the Edge. 

The observed possible \emph{negative} impact of some parameters combinations \emph{even with high initial accuracy} is a counterexample of the previously assumed unconditional inherent robustness. We believe that this is an interesting finding that can enable more comprehensively secure SNNs design. 
In future work, we will include deeper networks in our experiments, as we believe that the findings of this paper can be generalized to other SNNs and datasets. More complex behavior might be witnessed, but the very impact of structured parameters on SNNs robustness should be comparable.  %However, further investigations need to be done in this direction. Moreover, we believe that elaborating a theoretical model of robustness as a function of structural parameters is one of the promising orientations that might be very useful to the community, towards a more explainable behavior of SNNs.   

%\section{Conclusion}
%\input{conclusion}

\section*{Acknowledgment}

This work has been partially supported by the Doctoral College Resilient Embedded Systems which is run jointly by TU Wien's Faculty of Informatics and FH-Technikum Wien. This work is also partially supported by Intel Corporation through Gift funding for the project ``Cost-Effective Dependability for Deep Neural Networks and Spiking Neural Networks''.

\bibliographystyle{ieeetr}
{\rfsize \bibliography{main.bib}}
%\printbibliography

\end{document}